\documentclass{article}
\usepackage{graphicx} % Required for inserting images
\usepackage[backend=biber,sorting=none]{biblatex} % Bibliography management package
\usepackage{hyperref}
\usepackage{amsmath}
\usepackage{comment}
\usepackage{booktabs}
\usepackage[a4paper, margin=1.4in]{geometry}
\usepackage[normalem]{ulem}
\usepackage{xcolor}

\title{Anticipating Innovation Using Large Language Models}

\author{
    Enrico Maria Fenoaltea\textsuperscript{1, 2},
    Filippo Santoro\textsuperscript{2},\\
     Giordano De Marzo\textsuperscript{3},
     Segun Taofeek Aroyehun\textsuperscript{3},
    Andrea Tacchella\textsuperscript{2, 4}\\[1ex]
    \textsuperscript{1} Institute of Complex Systems (UBICS), Universitat de Barcelona\\
    \textsuperscript{2} Centro Ricerche Enrico Fermi (CREF),\\ Via Panisperna 89 A – 00184 Roma\\
    \textsuperscript{3} University of Konstanz  \\ 
    \textsuperscript{4} Complexity Science Hub (CSH),\\ Metternichgasse 8 1030 Vienna, Austria\\
    }

\date{}

\begin{document}

\maketitle
\begin{abstract}

Forecasting innovation, intended as the emergence of new technological combinations, is a fundamental challenge for science and policy. We show that forthcoming combinations leave an early trace in the collective language of patents, with predictive signals detectable even decades in advance. We show that signal is not attributable to any single inventor, but emerges as a collective shift in how technologies are described across thousands of patents. To this end, we introduce TechToken, a transformer-based model that treats technologies, classified by International Patent Classification codes, as words in its vocabulary, learning the language of technologies by embedding these codes during fine-tuning. We define context similarity between code embeddings as a measure of linguistic convergence and show that it accurately predicts first technological combinations. TechToken also improves general representation quality, outperforming state-of-the-art models across different patent-related tasks.

\end{abstract}

\section*{Introduction}

Predicting innovation presents a fundamental tension. Genuine innovations are unprecedented by definition, yet the ability to anticipate technological change carries significant implications for both corporate strategy and policy design \cite{eulaerts2026early,wipo2025transportation, oecd2023outlook}. Resolving this tension requires a theoretical framework that makes the space of possible innovations tractable.

Recombinant innovation theory provides a viable foundation to address this. Since Schumpeter first articulated that economic development proceeds through novel combinations of existing elements \cite{schumpeter1983theory, schumpeter2013capitalism}, a substantial body of research has established that a significant fraction of innovation develops as a combinatorial process \cite{youn2015invention, shi2023surprising}. Nowadays this process is largely understood in terms of gradual convergence of complementary and synergic technologies \cite{rosenberg1963technological, curran2011patent, sick2022exploring}. If innovation is recombination, the forecasting problem is transformed: rather than predicting the unprecedented, the task becomes predicting which groups of existing elements are most likely to combine \cite{youn2015invention}, by observing convergence patterns.

This reformulation naturally connects to the concept of the adjacent possible, developed by Kauffman \cite{kauffman2000investigations} and formalized in the context of innovation dynamics in terms of Polya's urns\cite{tria2014dynamics}. At any moment, the space of realizable innovations is bounded by what is reachable from the current configuration of knowledge. New combinations expand this frontier, opening further possibilities that were previously inaccessible. The adjacent possible is therefore not a fixed set but a dynamic boundary, and the central empirical challenge is identifying which combinations at this frontier are most likely to be actualized. Related work in economic complexity has shown that the structure of existing combinations (whether of technologies, products, or capabilities) carries substantial information about which new combinations are feasible \cite{hidalgo2009building, zaccaria2014taxonomy,tacchella2023relatedness}. Patent data are well suited to this task: patents systematically record recombination events, with each patent linking a set of technologies, represented by International Patent Classification (IPC) codes, into a documented combination. A first co-occurrence of two IPC codes in a patent marks the actualization of a previously unrealized adjacent possibility. While IPC codes can only provide a coarse grained description of a technology, they are good anchors to define the meaning and perimeter of a technological concept. Therefore, even if innovation can happen in nuanced ways, even \textit{within} an IPC class, using them as unit of analysis allows for a relatively simple and standardized framework to study the mechanism of technological convergence.

From a network science perspective, the task of predicting novel IPC combinations can be framed as a link prediction problem in the network of technological codes, where two codes are connected if they co-occur in at least one patent \cite{gao2017projection, baybusinov2024nonrandom}, and the goal is to predict the appearance of future links. Using topological information or bibliometric data from patents, numerous studies have adopted a combination of network science techniques \cite{liu2024prediction, hong2021effective, chang2025framework} and machine learning methods \cite{cho2021predicting, choi2022supervised, jiang2023dnformer, zhao2025forecasting} to forecast future convergence \cite{caviggioli2016technology}. %However, approaches relying solely on link information are inherently dependent on existing connections and may fail to capture diverse sources of convergence involving entirely new technologies \cite{passing2015measuring}.

The advances in Natural Language Processing (NLP) have enabled the integration of information extracted directly from patent texts into technology convergence prediction. For example, semantic analysis and topic modeling techniques have been used to extract technological topics from patents and to predict topic convergence in emerging fields \cite{afifuddin2024predictive, ma2021combining, liu2023technology} by looking at network topologies. Pre-trained transformer-based models \cite{vaswani2017attention}, such as sentence-transformers \cite{reimers2019sentence}, have been employed to generate vector representations (embeddings) of patents and scientific papers and to measure the relative distance between different topics \cite{afifuddin2025developing,feng2020proximity}.

%A growing literature has taken up the challenge of predicting these first co-occurrences, drawing on network science, machine learning, and natural language processing \cite{caviggioli2016technology, cho2021predicting, choi2022supervised, kim2020machine, tacchella2020language}. These approaches have steadily improved forecasting accuracy, but share a structural limitation: they treat the recorded history of combinations as the primary signal, extrapolating forward from what has already been connected. They are therefore blind to the earliest stages of an emerging combination--- precisely the moment when anticipation would be most valuable.

The central finding of this paper is that we can extract a predictive signal from the collective dynamics of language across the patent corpus that precedes the actualization of new combinations. Before two technologies are combined in a patent, the texts in which they separately appear begin to share semantic and lexical features. In other words, the linguistic contexts of two technologies converge years before their technological convergence is recorded. Crucially, this is not a signal that can be located in any single document or attributed to any individual inventor. It is an emergent property of the corpus as a whole, a distributed shift in how thousands of inventors, working independently, write about and around technological concepts. The predictive information is only visible at the level of the full population of patents, reflecting a collective dynamic in the evolution of technological discourse that precedes the actualization of new combinations by years, with statistical signatures observable already two decades in advance. This finding suggests that the adjacent possible leaves a legible trace in language before it is actualized is visible in technological data. In other words, that shifts in how communities of practice think and write about technologies are an early and observable signature of impending innovation.

Detecting this signal requires embeddings of technological codes that are genuinely grounded in linguistic context and sensitive to how the language surrounding a technology evolves across the corpus over time. Prior work establishes a related intuition: \cite{tacchella2020language} shows that technologies used in similar contexts are more likely to combine in the future, operationalizing context through Word2Vec embeddings of IPC code co-occurrence sequences. That approach, however, does not engage with the textual content of patents in a contextual way, as it operates on sequences of codes rather than language, thereby remaining closer in spirit to a network method, and produces static embeddings that are insensitive to how the meaning and use of a technology vary across patent contexts.

To generate rich contextual representation of technologies here we introduce TechToken, a transformer-based language model fine-tuned on patent text in which IPC codes are introduced directly as tokens in the model's vocabulary. During fine-tuning, the attention mechanism operates directly between the words of a patent abstract and the IPC code tokens associated with that patent, grounding the embedding of each technological code in the actual language used to describe the inventions in which it appears. Each code receives a distinct embedding for each patent context, capturing the genuine polysemy of technologies applied across diverse domains. Similarity between code embeddings, which we term context similarity (CS), serves as the forecasting signal. Importantly, CS is computed as an aggregate statistic over all embedding pairs from two codes across the full corpus, making it a global, emergent quantity rather than a property of any individual document.

We evaluate TechToken on out-of-sample patent data and show that it substantially outperforms all baseline models in predicting first co-occurrences of IPC codes, including considerably larger language models such as LLaMA 3.1. Performance improves as attention is restricted to the most statistically significant innovations (i.e. those with an unexpectedly high number of co-occurrences given the frequency of the IPC codes in the corpus), suggesting that the linguistic signal extracted by the TechToken model captures technological meaning rather than trends in patent volumes. Beyond innovation forecasting, TechToken achieves state-of-the-art performance on standard patent tasks including IPC classification, citation prediction, and title--abstract matching, confirming the quality of the embeddings it produces and suggesting that the addition of informative labels within the attention mechanism of transformers can significantly improve the quality of their internal representations.

The contribution of this paper is twofold. Empirically, we demonstrate that linguistic convergence precedes and predicts technological convergence, and that this signal is an emergent, corpus-level phenomenon reflecting the collective dynamics of technological discourse. Methodologically, we introduce a general framework for embedding symbolic classification codes directly into a language model's vocabulary, enabling the model to learn context-dependent representations of structured entities, an approach with potential applications beyond the patent domain.

%So, in technical term our task is to predict new combiantions of codes. Agreeing or not agreeing with this definition, this is challenging task for scientistist with interest form an economic and social policy and also from a technical point of of view as it can be expanded to many other context. The main idea is that the larger the similarity, in term of use case etc, between two technology the higher the probability that they will appear togheter. So one issue is how to measure this similarity. From a network science perspective bla bla, the the ML and AI tecnique emerged prepotentemente, and now there are LLMs, as we deal with text (patent), and we exploity the LLMs capabilities further by defining new vocabolary terms. Indeed, . bla bla bla (è molto simile a word to vec, quindi parlane. In sostanza, non solo inventiamo il linguaggio delkl'innovazione, ma lo ricostruiamo in relazione la linguaggio normale).

%Our contribution is twofold: 1) Innovation can be predicted better with LLM 2) new tech token model bla bla bla. Also we proov again that working on smaller models can beatr the brute firce of using as it is large models

%\section{Materials and Methods}

\section*{Results}
 We demonstrate quantitatively that the convergence of linguistic context is a systematic signal that anticipates visible technological convergence in terms of novel combinations of IPC codes. We show that the novel TechToken model that we introduce excels in extracting this signal, which is however visible also when other language models are used. Finally we demonstrate that the TechToken model produces state-of-the-art embeddings of patents that systematically outperform those obtained with comparable models available in the literature.

\subsection*{Innovation forecasting}
\begin{figure}[h!]
    \centering
    \includegraphics[width=1\textwidth]{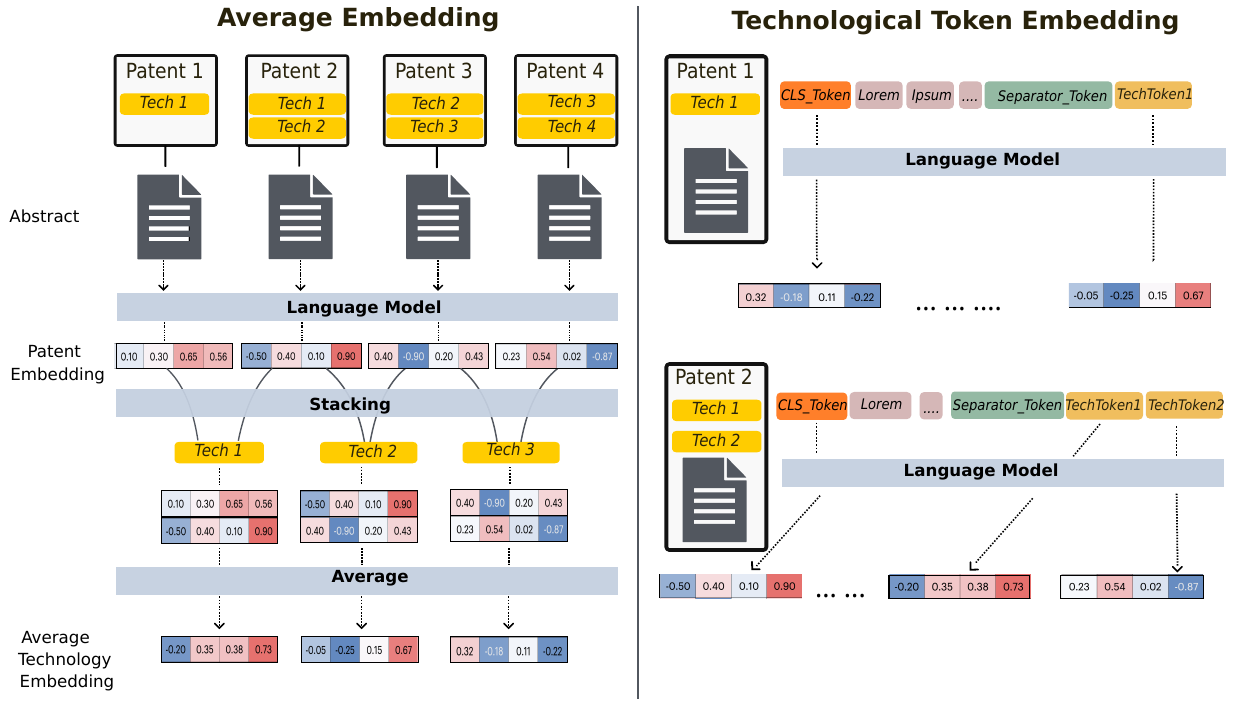} % Replace with your image file name
    \caption{\textbf{Scheme of the two technology embedding strategies}: In the average embedding strategy (left), patent embeddings are generated first using standard LMs, and the embeddings for an IPC code are computed as the average of all the embeddings of the patents it is associated with, resulting in each code having a unique embedding. In the Tech-Token embedding strategy (right), where we have expanded an LM's vocabulary by introducing new tokens that represent individual IPC codes (the technological tokens), embeddings are directly generated for the technological codes associated with patents, so that each code has multiple embeddings, one for each patent it belongs to.}
    \label{fig:embeddings} 
\end{figure}

Our approach relies on the idea that the linguistic contexts surrounding technologies start to converge earlier than the first combination of two technologies is observed. To quantify linguistic convergence we rely on the internal representations of Language Models (LM), i.e. their embedding vectors (see Methods). When processing a text, LMs transform them into vectors (embeddings) for them to be manipulated with mathematical operations. The topology of such embedding spaces capture semantic meaning: similar concepts tend to be associated with highly parallel vectors. Each word, or part of a word, is represented separately, and the embeddings are summed or averaged together to produce a single representation for the whole text. To compute embeddings for an IPC code, a straightforward approach is to collect all the embeddings of patents associated with such IPC and average them. This is the Average Embedding technique described in the left panel of Figure \ref{fig:embeddings} and is applicable to any LM that can generate an embedding for the patent text.

%In particular, we develop two strategies to obtain embeddings of IPC codes, schematically outlined in Fig.\ref{fig:embeddings}. The first approach is patent-based, meaning that we construct the embedding of an IPC code within a given time period as the average of the embeddings of all patents classified under that code. The patent embeddings are obtained in the standard way, using patent abstracts 
%as input to LLMs, either in their base form or fine-tuned as described in the Methods section.

%The second approach follows a different paradigm and is IPC code-based: rather than relying on patent embeddings, we extract embeddings directly for the IPC codes. To do this, we enhance the vocabulary of an LLM by adding new tokens corresponding to the technological codes associated with each patent. In a sense, fine-tuning teaches the model a new \textit{language} in which the letters are IPC codes, allowing the LLM to learn representations of these codes directly in its embedding space.

This approach, however, suffers from two significant shortcomings: i) all IPC codes present in a patent receive the same representation, even though they can represent very different technological concepts; ii) the attention mechanism, which is responsible for encoding the relations among words in Transformer-based language models, does not operate on the IPC codes, thereby ignoring the heterogeneity of their semantic relationship with each word in the patent's text and with the other IPC codes in the patent. To overcome these limitations we introduce the TechToken model, where IPC codes are treated as part of the patent's language. In particular, we develop a novel fine-tuning strategy (see Methods) to obtain embeddings of IPC codes by expanding the vocabulary of an LM with new tokens, the technological tokens, corresponding to the IPC codes associated with each patent. In a sense, fine-tuning teaches the LM a new \textit{language} in which the IPC codes are new words, allowing our TechToken model to learn representations of these codes directly in its embedding space. This approach seeks to leverage the functionality of the attention mechanism to build explicit attention links between the written language of the patent text and the technological tokens, as well as between the technological codes themselves. In other words, the strength of the TechToken method lies in the fact that the language model can build its representation of the new technology tokens in terms of the already known words used in the patent descriptions. Within this framework, as illustrated in the right panel of Figure \ref{fig:embeddings}, we can extract specific embedding vectors for each IPC code in each patent.

%Below, we compare the results obtained with TechToken to those obtained using a benchmark approach, the average embedding method, in which the embedding of an IPC code is constructed as the average of the embeddings of all patents classified under that code.
%Note that the average embedding method is patent-based: patent embeddings are obtained in the standard way by feeding patent abstracts into LLMs, either in their base form or after classical fine-tuning techniques, as described in the Methods section. TechToken instead follows a new paradigm, i.e., it is code-based: rather than relying on patent embeddings, it enables us to extract embeddings directly for IPC codes.  In the SI, we illustrate this point, i.e., how the model's understanding of the new tokens is grounded in the standard patent text using an explainable AI tool.

%The two methods by which a unique embedding for IPC codes can be obtained are schematically outlined in Figure~\ref{fig:embeddings} and detailed in the Methods section.

%In the following two subsections, we describe both strategies and the LLM architectures used in more detail.

%\subsection*{Innovation predictions}
Once constructed the embeddings for technological codes, we use them to quantify linguistic distance by defining context similarity (CS) between two technologies as the cosine similarity between their embeddings. An increase in CS signals that the language surrounding two IPC codes is getting is getting similar. In the case of the TechToken embeddings, we have multiple embeddings for each IPC code (one from every patent in the window considered). In this case, we compute the CS as the average similarity of the top 1\% couples of embeddings with the highest similarity. With this approach, we take care of the polysemanticity of the IPC codes (see SI): the same code can have different meaning in different contexts, which are captured by the TechToken embeddings. In our definition of CS we aim at assessing the distance between the two most similar meanings of two IPC classes\footnote{a more extreme approach would be to define CS as the highest cosine similarity among all the embeddings of two IPC classes, but such definition is noisy and we find that using a top-1 or top-10 \% approach provides better results.}. We now show that an increase in CS between two codes allows us to predict new combinations of technologies before they occur.

\begin{figure}[h!]
    \centering
    \includegraphics[width=\textwidth]{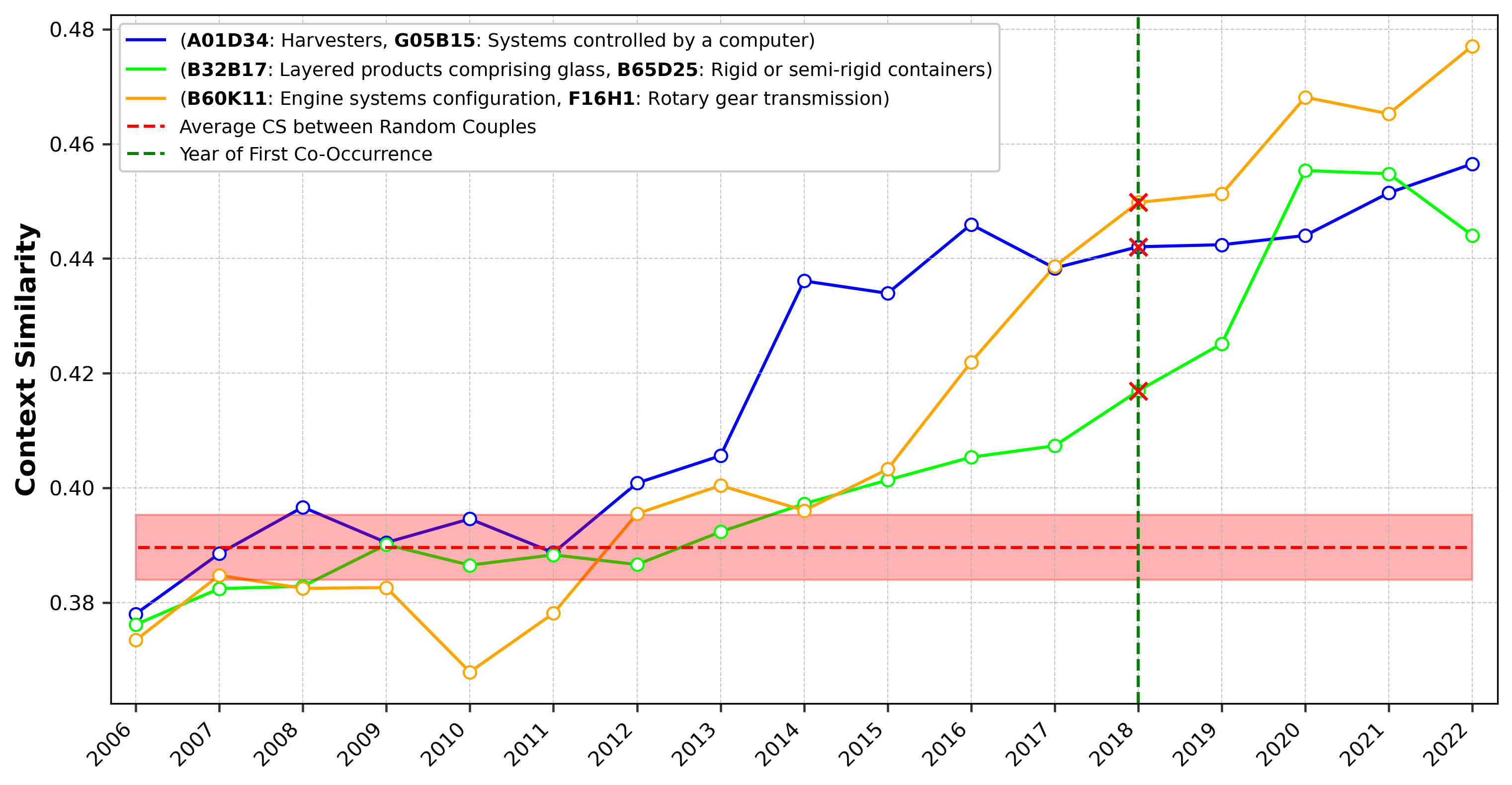}
    \caption{\textbf{CS vs. Time for three pairs of IPC codes.} For each couple of technologies, CS is computed over 1-year intervals as the average of the top $1\%$ highest cosine similarities among all possible pairs of token embeddings from the two IPC codes, generated using the TechToken method. Each point is the average of a three years sliding window. The red area represents three times the standard deviation of CS values calculated for random pairs of IPC codes. The vertical dashed line marks the year of the first co-occurrence of each of the three IPC code pairs.}
    \label{fig:pred} 
\end{figure}

We start by looking at a few cherry-picked examples of couples of IPC codes that fist co-occur in 2020. We compute their CS using patents in windows of 1 year and we compare it to the general distribution of CS among all IPC couples. The results are presented in Figure \ref{fig:pred}. Direct observation reveals that new co-occurrences are preceded by an increase in CS between the two codes. In Figure \ref{fig:pred}, we highlight three examples of code pairs from different sectors. It is clear that the linguistic signal of the first co-occurrence emerges years before the co-occurrence actually happens, specifically when the CS of the three pairs begins to rise and moves outside the bulk of the distribution. The same pattern can be observed even when using technology embeddings computed with the average embedding method.% This suggests that innovations are often foreshadowed by the convergence of the linguistic contexts in which the two codes typically appear.

This linguistic convergence occurs \textit{across} documents rather than \textit{within} documents, since, by construction, the code pairs are never observed together in the same patent before 2020. In other words, the CS signal we extract is a global property of the patent corpus and cannot be simply decomposed into the sum of individual contributions from single documents.

\begin{figure}[h!]
    \centering
    \includegraphics[width=1\textwidth]{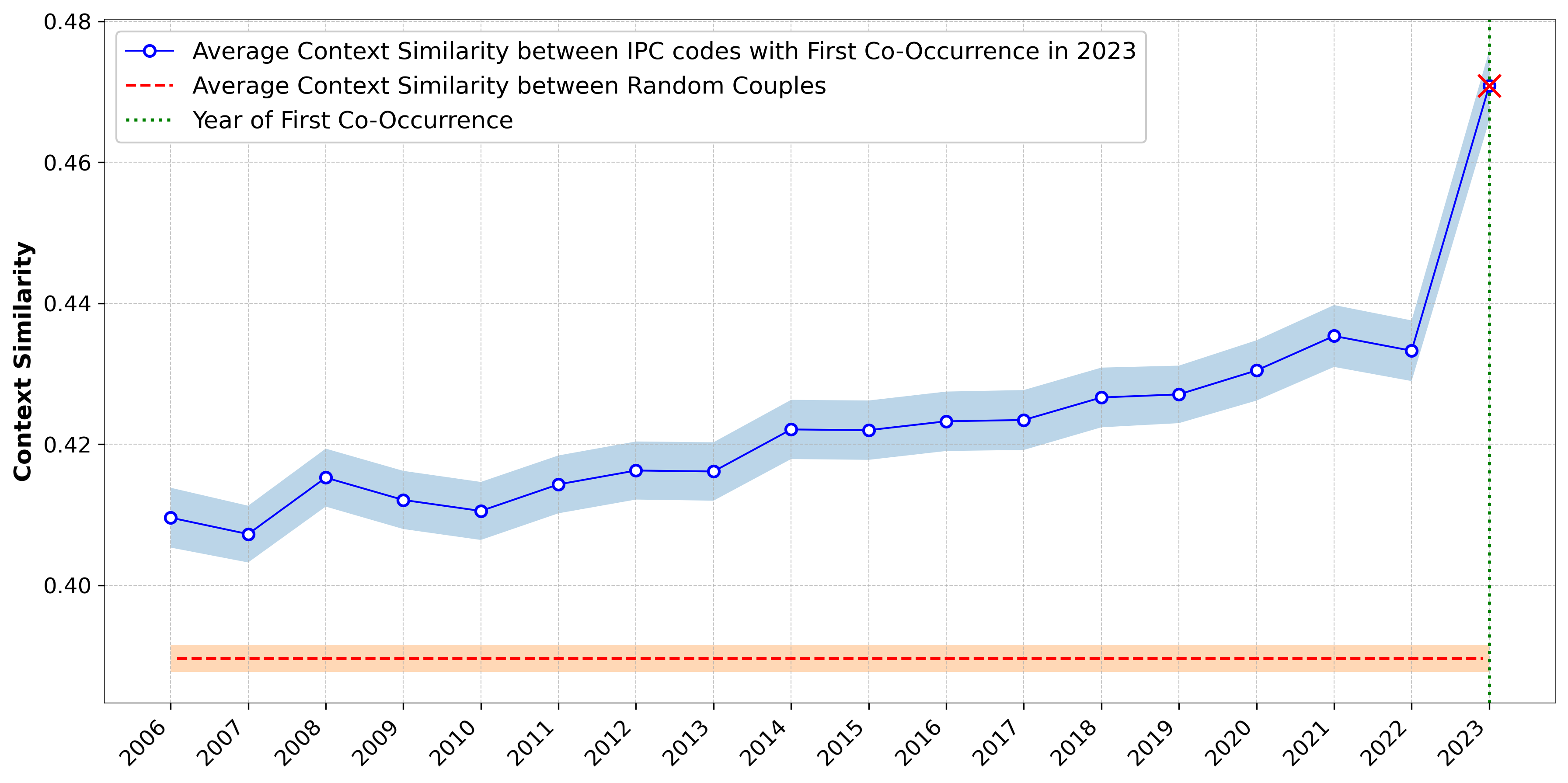} % Replace with your image file name
    \caption{\textbf{Average CS vs. Time.} For each couple of technologies, CS is computed over 1-year intervals as the average of the top $1\%$ highest cosine similarities among all possible pairs of token embeddings from the two IPC codes, generated using the TechToken method. The red dashed line represent the average CS between random pairs of codes and the shaded areas around the two lines represent their standard deviations.}
    \label{fig:pred_average} 
\end{figure}

Moreover, on average, the signal forecasting a first co-occurrence between a pair of technologies can be observed almost 20 years before the event itself. To show this, in Figure~\ref{fig:pred_average} we select all pairs of codes that had never appeared together in a patent before 2023 and compute their average CS in 1-year time windows from 2006 to 2023. Even in the earliest time window, the average CS is significantly above the confidence interval of the CS for a randomly selected pair of codes. Furthermore, the average CS of these pairs steadily increases until 2023, the year of their first co-occurrence, when the CS exhibits a sharp jump since, in that year, the contexts in which the two technologies are used are not only similar but identical (as the two codes appear together in at least one patent).

\begin{figure}[h!]
    \centering
    \includegraphics[width=1\textwidth]{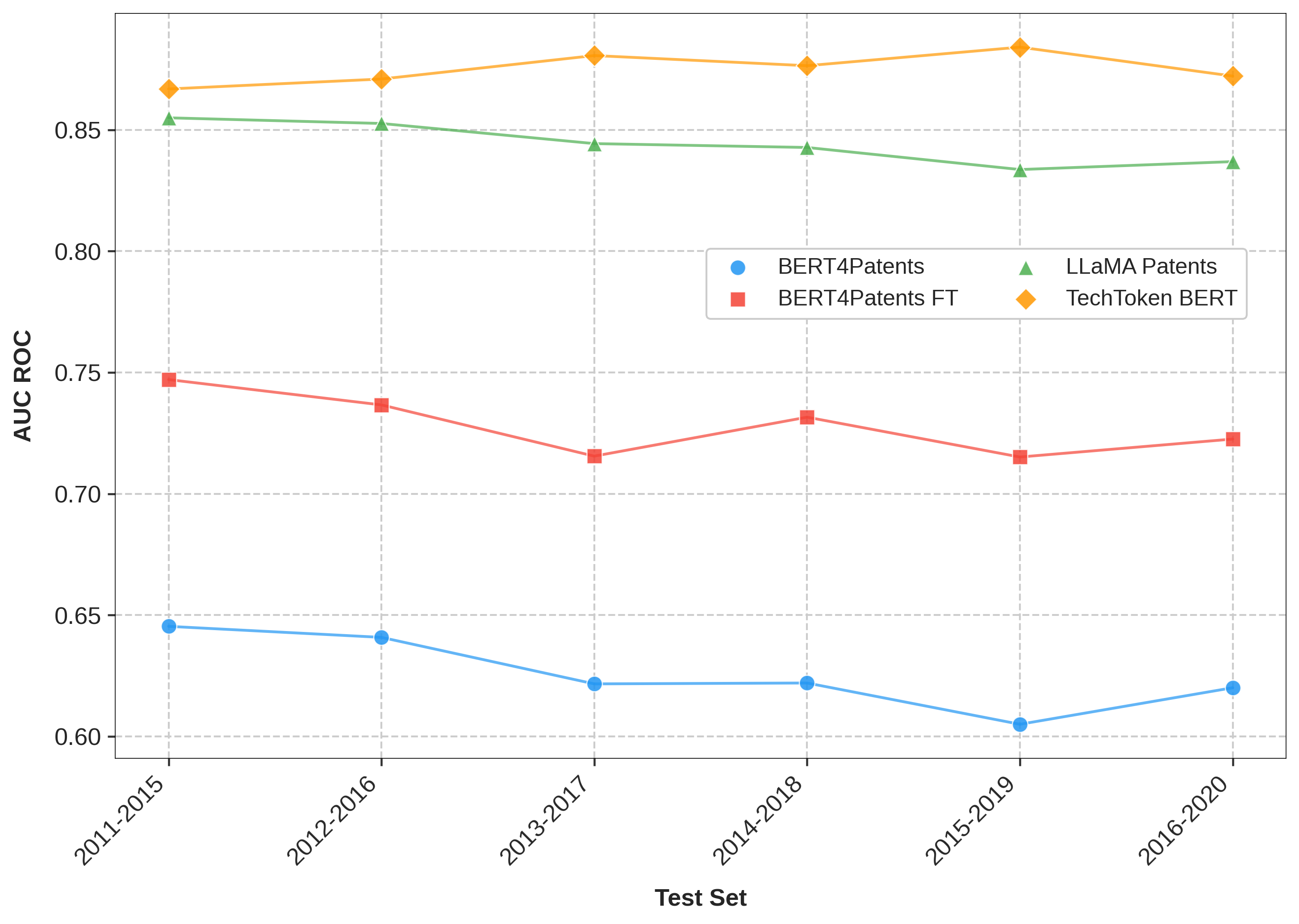} % Replace with your image file name
    \caption{\textbf{AUC-ROC vs. Time.} The metrics are computed using the CS values for pairs of codes from the 2006–2010 period that had never co-occurred before 2011, used as a classifier for the future time-window test sets. For each time window, we adjust the $z$-score threshold to maintain a fixed class imbalance of $0.005\%$. The embeddings of the IPC codes for the four curves are generated using the TechToken method with BERT (where CS is
computed as the average of the top $1\%$ highest CS among all possible pairs of token embeddings from the two IPC
codes-- see Methods) and the average embedding method, applied to the following models: BERT4Patents (base), a further fine-tuned version of BERT4Patents, and a fine-tuned version of LLaMA 3.1 8B.}
    \label{fig:ROC} 
\end{figure}

\begin{comment}
\begin{figure}[h!]
    \centering
    \includegraphics[width=1\textwidth]{Figures/fig_5a.png} % Replace with your image file name
    \caption{\textbf{Average CS vs. Time.} For each couple of technologies, CS is computed over 3-year intervals as the average of the top $10\%$ highest cosine similarities among all possible pairs of token embeddings from the two IPC codes, generated using the TechToken method.}
    \label{fig:pred_average} 
\end{figure}
\end{comment}
To systematically quantify the predictive power of our approach we set up a more thorough experiment. First we separate patents into a training and test set. We compute CS using patents from the training set for all IPC couples that had never co-occurred during or before the training set. We then record which of these couples do co-occur in the test set for the first time. Second we make the definition of what actually constitutes a significant new combination stricter, following~\cite{tacchella2020language}. Specifically we want to identify novel combinations that appear in the test set more frequently than expected by random chance and label only those as realized novel combinations. For each potential novel combination we compute the expected value and variance of the number of co-occurrences that would be expected if technological codes were randomly assigned to patents, while preserving the average number of connections for each patent and each technology (we use the Chung-Lu model \cite{chung2002connected}, see the Methods section for more details). For each couple we then compute the $z$-score as the difference between the observed and expected number of occurrences in the test set, divided by the square root of the variance (see Methods). This allows us to introduce a threshold on how strict we want to be on the statistical definition of what constitutes a real novel combination and what we can consider noise. More precisely, we use the $z$-score as a ranking and assign the $K$ couples of IPC codes with the highest observed $z$-score to the label $1$ (realized innovations) and the remaining couples to the label $0$. By having a fixed number of true positive examples we also fix the class imbalance ratio, allowing easier and more robust comparison of the results across different years. In the rest of the manuscript we report results at different class-imbalance ratio (CI) levels.

We then test the effectiveness of the CS metric as a classifier. We compute the CS of code pairs that had never co-occurred until 2010, using patents from 2006 to 2010. Finally, we evaluate how well this metric predicts the class these pairs belong to in future time intervals, starting with the 2011–2015 period and progressively shifting the 5-year window forward by one year. We use a fixed CI of $0.005\%$ (see Methods for the rationale behind this choice).

In Figure~\ref{fig:ROC}, we present the Area Under the ROC Curve (AUC) obtained using different LMs and embedding methods. The three BERT models are all based on the Bert for Patents (a plain version, a further fine-tuned version with the Mirror Bert approach and the TechToken version). The Llama model is a version of Llama 3.1 8B, fine-tuned of patent data with the LLM2Vec approach. See the method section for details on all the models and fine-tuning strategies. The figure shows that CS calculated with the TechToken method significantly outperforms all other classifiers across all future time windows and, unlike the others, maintains stable or slightly increasing  performance over time. This suggests that training a language model to explicitly represent IPC codes in its latent space, by introducing an additional vocabulary composed of IPC code tokens, substantially improves the model’s ability to understand the contexts in which technologies are used. Notably, with this technique we were able to boost the performance of a relatively small model such as BERT, enabling it to outperform much larger models such as LLaMA 3.1 8B, which is approximately 25 times lager in terms of parameters count.

%To gain an intuition for the predictive performance of the TechToken method, consider the top 1000 IPC code pairs in the 2011-2015 test set with the highest CS values computed in the 2006-2010 train set. With the strict class imbalance threshold fixed at $0.005\%$, approximately 40 of these pairs are true positives (i.e., actual innovations according to our definition). This means our method performs about 800 times better than random guessing (as a random set of 1000 IPC couples contains an expected value of 0.05 true innovations).

\begin{table}[h!]
\centering
\resizebox{\textwidth}{!}{
\begin{tabular}{|c|c|c|c|c|}
\hline
\textbf{MODEL}& AUC-ROC ($0.005\%$) & AUC-ROC ($0.004\%$) & AUC-ROC ($0.003\%$) & AUC-ROC ($0.0025\%$) \\
\hline
\textbf{BERT4Patents} & 0.725&0.716 &0.693 & 0.682  \\
\hline
\textbf{BERT4Patents FT} &0.765& 0.761 & 0.753 & 0.745 \\
\hline
\textbf{LLaMA Patents}& 0.856 &0.832  & 0.846 & 0.840 \\
\hline
\textbf{TechToken BERT (IPC)} & \textbf{0.936} & \textbf{0.938} & \textbf{0.940} & \textbf{0.940} \\
\hline
\textbf{TechToken BERT (CLS)} &0.908 &0.911  & 0.903 & 0.904 \\
\hline
%\textbf{TechToken BERT (CLS+IPC)} & 0.938  & 0.939 & 0.940 \\
%\hline
\end{tabular}
}
\caption{\textbf{AUC-ROC scores for different models and class-imbalance ratios.} 
The values in parentheses indicate the proportion of positive examples in the test set (class imbalance). All IPC code embeddings, except for the TechToken models involving technological token embeddings, are computed using the average embedding method. Each row reports the best score for each model, selected from the results obtained using mean pooling and those obtained using the CLS token embedding. The TechToken method is evaluated in two variants: i) Average of closest technological tokens, where CS is computed from the average of highest cosine similarities among all token-embedding pairs from the two codes; ii) TechToken CLS, where IPC code embeddings are computed using the average embedding method using the CLS token from the TechToken model.% and (iii) a combination of the CLS and IPC code token embeddings. 
Results are based on code pairs from 2019–2023 that had never co-occurred before 2024, with the test set consisting of patents published in 2024.}
\label{tab:auc_roc}
\end{table}

These results, however, are not strictly out of sample. The Llama 3.1 model has been trained with a knowledge cutoff date at December 2023\footnote{\url{https://github.com/meta-llama/llama-models/blob/main/models/llama3_1/MODEL_CARD.md} - Accessed April 2026}, and we fine-tuned the BERT models with patents up until that date. To provide a stricter validation, in Table~\ref{tab:auc_roc} we present an additional comparison of the predictive performance of the same models of Figure~\ref{fig:ROC} with a different test set. In particular, we calculate CS for code pairs that had never co-occurred before 2024, using the 2019–2023 period, and evaluate on patents published in 2024. This setup also guarantees that none of the tested models has been fine-tuned on the test set patents, as all models considered in this work were released before 2024 and fine-tuned with patents released in 2023 at the latest. For this comparison, we show AUC-ROC values for different class-imbalance ratios. To further disentagle the contribution of the embedding method (TechToken versus average embedding) from the mere quality of the models, we also report the TechToken CLS variant, where IPC code embeddings are computed using the average embedding method, with the CLS token from the TechToken model used as the patent embedding.%Since in this case the test spans only one year, we report the results for lower class imbalance ratios, where only highly statistically significant code combinations are classified as innovations. 

%; and (ii) a combination of the CLS and IPC code token embeddings. In particular, the CS is computed as the sum of the CS obtained using only the IPC code tokens and the CS obtained using only the CLS token. %Top-$10\%$ average version, where CS is computed as the average of the top $10\%$ highest cosine similarities among all possible pairs of token embeddings from the two IPC codes; ii) full-average version, where CS is computed as the cosine similarity between the embeddings of the two IPC codes, where each code’s embedding is the average of all its associated token embeddings.

%The Top-$10\%$ average version of the TechToken model 

The TechToken-based approaches significantly outperform all other models, regardless of the embedding technique (IPC or CLS). The IPC approach outperforms the average embedding based on CLS within the TechToken class of models. Besides, note that a higher class-imbalance ratio corresponds to treating as innovations those code pairs with lower $z$-scores, i.e., being less restrictive in the definition of innovation. The fact that our TechToken (IPC)method improves its relative performance as the class imbalance decreases suggests that this approach is especially effective in predicting the most significant innovations. See the SI Figure 5 for the full analysis on class imbalance the models' performances with the class imbalance ratio.
%Furthermore, the superior performance of the Top-$10\%$ version over the full-average version indicates that, in calculating CS, focusing only on the most similar token embeddings can reduce noise and enhance predictive accuracy.

\subsection*{Other patent-related tasks}
To understand if the performance in the innovation-prediction task is the reflection of a generally improved understanding of the technology language, we tested the performance of our model as a patent embedding model on three classical patent-related tasks \cite{jiang2025natural} and compared it with the best models available in the literature for these tasks. See the methods section for references of the models we compared. The tasks considered are: (i) IPC classification, i.e., assigning the correct IPC codes to a given patent. In this case, for our TechToken model, classification is performed by providing the patent title and abstract as input and using next-token prediction probabilities, restricted to the IPC code tokens. All codes whose predicted probability exceeds a chosen threshold (optimized out of sample) are considered to be assigned to the patent by the model(see the SI). For the other models, IPC prediction is performed using a k-nearest neighbor approach introduced in \cite{bekamiri2024patentsberta}, where a patent is assigned the IPC codes of the top-k most similar patents, based on cosine similarity between embeddings.   %where the threshold is chosen to maximize the F1 score on a validation set (see the SI for further details). %assigning all codes whose probability exceeds an optimal threshold. 
(ii) Citation prediction, i.e., the task of correctly predicting, for a given patent, which patents it cites. Here, we generate patent embeddings (we use the CLS token for the TechToken model) and employ embedding similarity to predict citations. This task is performed on the dataset provided by the authors of~\cite{ghosh2024paecter}\footnote{data is available at \url{https://huggingface.co/mpi-inno-comp/paecter} - Accessed April 2026} and with the same methodology. For this task we fine-tune an additional version of TechToken, i.e. we use TechToken as a base model and we perform the same fine-tuning described in~\cite{ghosh2024paecter} (see SI). (iii) Title–Abstract matching, i.e., the task of correctly pairing a patent abstract with its corresponding title. In this case as well, different LMs are used to generate abstract and title embeddings, which are then compared based on their similarity (see SI for details).

In all three tasks, TechToken achieves the best performance metrics, surpassing state-of-the-art embedding models specifically designed for these tasks, such as PatentSBERTa for code classification \cite{bekamiri2024patentsberta} and Paecter for citation prediction \cite{ghosh2024paecter}, demonstrating that the TechToken technique provides a strong foundational model that can be applied to a variety of tasks. Table~\ref{tab:patent_tasks} summarizes the results for the three tasks, reporting different performance metric for each task (see SI for a full description). TechToken is the best model in all task and according to all metrics, except for the MRR metric in the Title-Abstract matching task, where PatentSBERTa achieves an higher score. A more extended set of results is available in the SI, where we compare different embedding approaches (pooling and CLS) when they are applicable. Here we only report the best result for each model.

\begin{table}[ht]
\centering
\resizebox{\textwidth}{!}{%
\begin{tabular}{l | ccc | ccc | cc}
\toprule
 & \multicolumn{3}{c|}{IPC classification} & \multicolumn{3}{c|}{Citation prediction} & \multicolumn{2}{c}{Title--abstract matching} \\
Model & Macro F1\,$\uparrow$ & Micro F1\,$\uparrow$ & Ham.\ loss\,$\downarrow$ & RFR\,$\downarrow$ & MAP\,$\uparrow$ & MRR@10\,$\uparrow$ & AUC-ROC\,$\uparrow$ & MRR\,$\uparrow$ \\
\midrule
BERT4Patents        & 0.354 & 0.301 & 0.0012 & 1.57 & 59.46 & 80.60 & 0.920 & 0.799 \\
BERT4Patents FT     & 0.262 & 0.225 & 0.0015 & 1.88 & 52.78 & 73.16 & 0.832 & 0.427 \\
PatentSBERTa        & 0.356 & 0.303 & 0.0012 & 1.79 & 55.89 & 75.95 & 0.985 & \textbf{0.912} \\
Paecter             & 0.420 & 0.378 & 0.0010 & 1.32 & 68.11 & 88.14 & 0.944 & 0.896 \\
%LLaMA Wiki          & 0.347 & 0.296 & 0.0012 & ---  & ---   & ---   & 0.930 & 0.792 \\
LLaMA Patents       & 0.343 & 0.295 & 0.0012 & 1.74 & 56.78 & 75.47 & 0.973 & 0.766 \\
\midrule
TechToken BERT      & \textbf{0.488} & \textbf{0.449} & \textbf{0.0002} & 1.59 & 59.73 & 79.70 & \textbf{0.994} & 0.903 \\
TechToken BERT$^\dagger$ & --- & --- & --- & \textbf{1.26} & \textbf{68.96} & \textbf{89.54} & --- & --- \\
\bottomrule
\end{tabular}%
}
\caption{\textbf{Performance on three patent-related tasks.} Best score per column is marked in bold.
  Each model's score is the best across mean-pool and CLS embedding variants.
  TechToken IPC classification uses next-token probability thresholding rather than
  top-$k$ cosine similarity. $^\dagger$Fine-tuned on the Paecter citation dataset;
  reported only for the citation task.}% LLaMA Wiki was not evaluated on citation prediction.}
\label{tab:patent_tasks}
\end{table}

\begin{comment}
\begin{table}[h!]
\centering
\resizebox{\textwidth}{!}{
\begin{tabular}{|c|c|c|c|}
\hline
\textbf{MODEL} & RFR & MRR & MAP \\
\hline
\textbf{BERT4Patents} & 2.324 & 0.757 & 0.532 \\
\hline
\textbf{BERT4Patents FT} & 2.140 & 0.738 & 0.486 \\
\hline
\textbf{PatentSBERTa} & 2.172 & 0.760 & 0.517 \\
\hline
\textbf{Paecter} & \textbf{1.503} & \textbf{0.882} & \textbf{0.682} \\
\hline
\textbf{LLaMA Wiki} & 1.973 & 0.805 & 0.566 \\
\hline
\textbf{LLaMA Patents} & 1.793 & 0.816 & 0.583 \\
\hline
\textbf{Tech-Token BERT (CLS)} & 2.101 & 0.777 & 0.556 \\
\hline
\textbf{Tech-Token BERT (Mean Pool)} & 2.435 & 0.733 & 0.531 \\
\hline
\end{tabular}
}
\caption{}
\label{tab:your_label}
\end{table}
\end{comment}

\section*{Conclusion}

Innovation is largely understood as a process of convergence and recombination, in which previously separate technological or scientific ideas are assembled into novel combinations. This process has been characterized, largely ex post, as the gradual expansion of an adjacent possible, the evolving frontier of combinations reachable from the current configuration of knowledge. Our central finding is that these processes are anticipated by visible, quantifiable signals in the textual record of innovation itself. Before two technologies are first combined in a patent, the linguistic contexts in which they separately appear begin to converge, with statistical signatures visible already two decades in advance. The adjacent possible, in other words, is not silent. It leaves a legible trace in the collective discourse of inventors well before it is realized as a documented co-occurrence of IPC codes.

This trace is emergent in a strict sense. In general it cannot be located in any individual patent, attributed to any single inventor, or reconstructed from any one document's content. It exists only at the level of the corpus as a whole, as a distributed shift in how thousands of independent actors write about technologies that have yet to meet. In this respect, our results give a concrete empirical anchor to an idea that has long animated the literature on recombinant innovation and economic complexity: that the feasibility of a new combination is encoded in the structure of what already exists, and that communities of practice gradually align the conceptual territory around technologies before formally bridging them. Where prior work has operationalized this intuition through co-occurrence networks or static embeddings of technological codes, we show that the signal is richer, earlier, and more linguistically grounded than these approaches can capture. Technological convergence is preceded by semantic convergence, and semantic convergence is a measurable, corpus-level phenomenon.

Making this signal visible required a methodological step whose implications extend beyond patents. The TechToken architecture treats IPC codes as tokens in the language model's vocabulary and learns their embeddings jointly with patent text during fine-tuning, binding each code to the evolving language used to describe the inventions in which it appears. This design allows the attention mechanism to operate directly between natural-language tokens and symbolic classification tokens, producing context-dependent representations that capture the polysemy of technologies deployed across heterogeneous domains. Empirically, the approach enables a BERT-scale model to substantially outperform a model roughly twenty-five times larger on the forecasting task, and simultaneously achieve state-of-the-art results on IPC classification, citation prediction, and title–abstract matching. More generally, embedding structured symbolic labels into a language model's vocabulary is a broadly applicable recipe: any domain in which documents are tagged with codes from a controlled vocabulary (e.g. scientific publications and MeSH or arXiv categories, products and harmonized trade classifications) admits the same treatment.

Several limitations should be acknowledged, and each suggests a natural extension. Our analysis operates at the IPC group level, which provides a standardized and tractable unit but inevitably coarse-grains the technological space: genuine recombination within a group is invisible to our framework. Relatedly, we have restricted attention to pairwise combinations, whereas innovation frequently involves higher-order assemblies of three or more technologies whose joint actualization cannot be reduced to their pairwise precursors. Both limitations can be addressed within the same methodological logic and can be implemented as natural extensions in future works. Because TechToken grounds IPC embeddings in the specific language surrounding each patent, the framework opens a path toward tracking convergence at finer semantic resolution (e.g. between keywords, technical terms, or extracted concepts) and toward modeling the simultaneous alignment of multiple technological vocabularies as a multi-way phenomenon. Pursuing these directions would move the study of the adjacent possible from the detection of pending bilateral combinations toward a more complete cartography of the technological frontier.

Taken together, our results argue that the collective language of invention is a leading indicator of technological change, and that relatively lightweight language models, when equipped with the right symbolic scaffolding, can read that indicator with precision. For innovation policy, corporate R\&D strategy, and any domain in which anticipating recombination carries real stakes, this offers something that has long been promised but rarely delivered: a principled, early, and empirically grounded window onto what is about to become possible.

\section*{Data and Methods}
\subsection*{Data}
In this study, we rely on data from the full European Patent Bulletin AB, which contains information on over one million patents, including publication dates, abstracts, claims, and associated technologies classified according to the International Patent Classification (IPC) system. The IPC is a hierarchical system used to categorize patents based on the technical content of their inventions. Each patent is assigned one or more IPC codes that describe the technological areas it pertains to. These codes follow a structured alphanumeric format that reflects the underlying technology. The system divides all fields of technology into eight main sections, which are further subdivided into classes, subclasses, groups, and subgroups. The IPC is regularly updated to reflect new and evolving technologies.

In our analysis, we consider IPC codes truncated at the group level (excluding subgroups), resulting in a set of 7200 unique codes for patents spanning the period from 1980 to 2024. The reason for truncating is simply to reduce computational complexity; however, our approach is applicable at all levels of aggregation.

For the LLMs fine tuning, described in detail below, we consider patents published between 1980 and 2005. To test our prediction model, we use patents published between 2006 and 2024, during which the IPC classification is consistent with the 8th version of the system. We focus exclusively on patents written in English, resulting in a final dataset of 1302500 patents. Patents missing either abstracts or claims have been excluded from the analysis.

\subsection*{Methods}

\subsubsection*{TechToken model}

The central idea of the TechToken model is to treat IPC codes as part of the patent's language. Rather than representing a code through the aggregate embeddings of the patents it labels (i.e. the approach taken by the average embedding method described in the next section) we extend the vocabulary of a pre-trained language model, BERT \cite{devlin-etal-2019-bert} in our case, with one dedicated token for each IPC code.

Fine-tuning is performed in a standard masked-language-modelling setting on sequences of the form:
$$
\texttt{[CLS]}\ \textit{patent title}\ \texttt{[SEP]}\ \textit{patent abstract}\ \texttt{[SEP]}\ \texttt{[TT}_1\texttt{]}\ \texttt{[TT}_2\texttt{]}\ \ldots\ \texttt{[TT}_N\texttt{]}
$$
where each $\texttt{TT}_i$ (TechToken) corresponds to one of the IPC codes associated with the patent. Through this procedure, the attention mechanism learns to link each technological token both to the words of the abstract and to the other technological tokens in the same patent. In effect, fine-tuning teaches the model a new language in which IPC codes are words, and their meaning is grounded in the linguistic contexts in which they are used. Because each patent provides a distinct context, each IPC code acquires a distinct embedding for every patent in which it appears, naturally capturing the polysemy of technologies deployed across heterogeneous domains.

At inference time, we recover the embedding of an IPC code in a given patent by reading the corresponding technological-token vector from the model's last hidden layer. This yields a set of context-specific embeddings for each code, one per patent, which can be used directly, or aggregated (e.g., by averaging or by selecting the top-similarity pairs, as in our definition of context similarity) to obtain a single representation of the code over a chosen time window.

The TechToken model can also be used as a general-purpose patent embedding model. In this case, we use the \texttt{[CLS]}\footnote{The \texttt{[CLS]} token is a special token prepended to every input sequence in BERT-style models. During training, its final-layer embedding is optimized to aggregate information from the entire sequence through the attention mechanism, and is conventionally used as a single-vector representation of the full input for downstream tasks such as classification or similarity computation.} token embedding as the patent representation, and construct the embedding of an IPC code as the average of the \texttt{[CLS]} embeddings of all patents labelled with that code. That is, the average embedding (see next section) method applied to TechToken. The advantage over a standard fine-tuned language model is that the \texttt{[CLS]} token now encodes information jointly from the patent text \emph{and} from the combination of technological codes assigned to the patent, since both have participated in the attention computation during fine-tuning. As shown in the main text and further documented in the Supplementary Information, \texttt{[CLS]}-based TechToken embeddings are themselves highly effective for IPC combination prediction, and achieve state-of-the-art performance on standard patent tasks.

\subsubsection*{Average embeddings method}
The average embedding method is our benchmark method to validate the advantage of TechToken. Since this approach is patent-based, the first step is to generate domain-specific representations (embeddings) of patent text. To this end, we use BERT4Patent and Llama3.1 (8B). We fine-tune these models with standard techniques to better capture the linguistic characteristics of patent texts, which often feature specialized terminology and formal language (see the Methods section). We aim to optimize their performance in generating context-aware embeddings that accurately reflect the semantic content of the input.

For both BERT4Patent \cite{srebrovic2020leveraging} and Llama3.1 8B \cite{grattafiori2024llama}, we generate contextual embeddings for each token in the input text (the patent abstracts). After obtaining these embeddings, we average the token-level embeddings across the entire document to create a single vector that captures the overall semantic content of the patent. The embeddings can be used for various downstream tasks such as classification, clustering, or similarity detection. 

Finally, the embedding for an IPC code in a given time period is calculated as the average of the embeddings of all patents linked to that code in that time interval. %In the Supplementary Information (SI) we provide a qualitative validation of the embeddings we produce by presenting a two-dimensional representation of the embedding space we generated for patents and technologies.

\subsubsection*{Models and fine-tuning strategies}
For the average embedding approach, we fine-tuned BERT4Patent and Llama3.1. 

BERT4Patent is a variant of the BERT architecture, pre-trained on patent data to better capture the nuances and vocabulary specific to patent documents. The fine tuning is performed using patent data from 1980 to 2005 in order to keep some out-of-sample data (from 2006 to 2024) to test the accuracy of the prediction model. BERT’s transformer architecture is highly effective for generating contextual embeddings for each token, making it well-suited for tasks requiring a deep understanding of text semantics. To fine-tune BERT4Patent, we use the Mirror-BERT approach \cite{liu2021fast}, which focuses on improving sequence-level representations through unsupervised contrastive learning. This allows the model to generate embeddings that better reflect the overall context and content of patent texts.

Llama3.1 (8B) is an LLM with capacity to support long textual sequences. Although Llama3.1 is not specifically trained on patent data, we fine-tune it on domain-specific patent texts to produce high-quality contextual embeddings, which can be applied to longer documents. For the fine-tuning process, we leverage the LLM2Vec approach \cite{behnamghader2024llm2vec}, which adjusts the architecture to make it bidirectional and introduces two further training steps: masked next token prediction and unsupervised contrastive learning. We apply parameter-efficient training using LoRA (Low-Rank Adaptation), ensuring the model adapts effectively and efficiently to the patent domain.

Finally, to benchmark our TechToken approach on additional patent-related tasks, we use PatentSBERTa \cite{bekamiri2024patentsberta} and Paecter \cite{ghosh2024paecter}, particularly for IPC classification and for citation prediction, respectively.

\subsubsection*{Context similarity for innovation prediction}
Once we have obtained the embeddings for patents and technological codes, we use them to predict innovations. As mentioned above, the core idea of our prediction framework is that two technologies frequently applied in similar contexts are highly likely to be used together in the same patent. It is then natural to define context similarity (CS) between two technologies as the overlap between their embeddings. Our key finding is that an increase in CS between two codes allows us to predict new combinations of technologies before they occur.

Technically, with the average embedding method, where the embedding of an IPC code is computed as the average of the embeddings of the patents classified under that code, the CS between two technologies is simply the cosine similarity between their respective embeddings.
When using the TechToken model, each technology is represented more explicitly by the different embeddings corresponding to the tokens related to that IPC code across different patents. Hence, the embedding of a technology can still be defined as the average of all its embeddings, and, again, the CS between two IPC codes is the cosine similarity between their embeddings.
However, we further improve prediction performance by computing the CS between two technologies as the average of the top $x\%$ (we use $x=1\%$ in the main text) highest cosine similarities among all possible pairs of token embeddings from the two IPC codes. Restricting  to the closest technological-token pairs reduces bias arising from technologies whose codes are used in very heterogeneous contexts and thus, whose token embeddings span a large region of the embedding space. Focusing on the top 
$x\%$ of similarities therefore isolates the most relevant portions of the embedding space when computing technological proximity.

\subsubsection*{A stricter definition of innovation}\label{innovation_definition}

As mentioned in the introduction, the type of innovation we aim to predict is defined as a never-seen-before association of preexisting elements. Formally, the simplest definition of innovation in our context is the first co-occurrence of two IPC codes in a patent, i.e., an unprecedented association of two technologies.

However, in this study, we adopt a stronger definition, already used in \cite{tacchella2020language}, to ensure that the new combination of technologies represents a genuinely innovative and persistently used association, rather than a spurious co-occurrence driven, for example, by the popularity of one of the two codes. In particular, we define a new association of codes as innovative if the observed number of co-occurrences between technologies in patents, within a given time window, is significantly higher than expected under a null model of random bipartite graphs connecting patents and technologies \cite{chung2002connected, saracco2017inferring}.

The null model predicts that the probability $P_{pc}$ that patent $p$ is connected to technology $c$ is proportional to the product of how many technologies patent $p$ has, and in how many patents technology $c$ is found. Formally, we assume:
$$P_{pc}=\frac{w_p w_c}{N}, $$
where $w_c$ is the degree of the technology code $c$, $w_p$ is the degree of the patent $p$, and $N$ is total number of links in the patents-technologies bipartite network. From this, we calculate a z-score for each pair of codes $c$ and $c'$ as follows:
$$Z_{cc'}=\frac{O_{cc'}-E_{cc'}}{\sigma_{cc'}},$$
where $O_{cc'}$ is the observed co-occurrence value in the considered time window for the couple $cc'$, $E_{cc'}$ is its expected value computed as $E_{cc'}=\sum_p P_{pc}P_{pc'}$, and $\sigma_{cc'}$ is the standard deviation calculated as $\sigma_{cc'}=\sqrt{\sum_p P_{pc}P_{pc'}(1-P_{pc}P_{pc'})}$. 

Hence, we define a new couple of codes $cc'$ as an innovation only if its z-score exceeds a specified threshold. This threshold quantifies how unexpected the success of a technology pair is, given the degree sequences in the patents-technology bipartite network. In this study, we evaluate the predictive performance of our model by selecting a z-score threshold for each test set to maintain a consistent class imbalance ratio between positive and negative examples.
This approach is particularly convenient because the AUC-ROC metric, which we use to evaluate performance, is highly sensitive to class imbalance.

The choice of a very small class imbalance ratio in Fig.~\ref{fig:ROC} and Table~\ref{tab:auc_roc} is motivated by statistical considerations. Given the null model described above, we compute the $p$-value associated with an observed number of co-occurrences for each pair of codes, corresponding to a given $z$-score. Note that the distribution of co-occurrences follows a Poisson binomial distribution.
We therefore select a $z$-score threshold such that the number of code pairs classified as real innovations is significantly larger than the expected number of false positives, given by $p\text{-value} \times N$, where $N$ is the number of potential code pairs. For example, in the 2011–2015 test set, there are approximately $2\times10^{7}$ possible code pairs. A class imbalance of $0.005\%$ corresponds to $\sim10^{3}$ positive instances. We therefore choose a $z$-score corresponding to a $p$-value of order $10^{-5}$, ensuring that the identified innovations are highly unlikely to arise from random noise.

\printbibliography

\end{document}